\documentclass[conference]{IEEEtran}
\IEEEoverridecommandlockouts
\usepackage{cite}
\usepackage{url}
\usepackage{makecell}
\usepackage{multirow} 
\usepackage{amsmath,amssymb,amsfonts}
\usepackage{algorithmic}
\usepackage{graphicx}
\usepackage{textcomp}
\usepackage{xcolor}
\usepackage{epstopdf}
\usepackage{svg}
\def\BibTeX{{\rm B\kern-.05em{\sc i\kern-.025em b}\kern-.08em
    T\kern-.1667em\lower.7ex\hbox{E}\kern-.125emX}}
\begin{document}

\title{SemIRNet: A Semantic Irony Recognition Network for Multimodal Sarcasm Detection
}

\author{
\IEEEauthorblockN{1\textsuperscript{st} Jingxuan Zhou $\dagger$}
\IEEEauthorblockA{
\textit{University of New South Wales}\\
Canberra, Australia \\
zhou20040626@outlook.com}
\and
\IEEEauthorblockN{2\textsuperscript{nd} Yuehao Wu $\dagger$ *}
\IEEEauthorblockA{
\textit{University of Sydney}\\
Sydney, Australia \\
yuwu6640@uni.sydney.edu.au}

\and
\IEEEauthorblockN{3\textsuperscript{rd}Yibo Zhang}
\IEEEauthorblockA{
\textit{University of New South Wales}\\
Canberra, Australia \\
bernie.zhangyibo@gmail.com}

\and
\IEEEauthorblockN{4\textsuperscript{th} Yeyubei Zhang}
\IEEEauthorblockA{
\textit{University of Pennsylvania}\\
Philadelphia, United States\\
joycezh@alumni.upenn.edu}

\and
\IEEEauthorblockN{5\textsuperscript{th} Yunchong Liu}
\IEEEauthorblockA{
\textit{University of Pennsylvania}\\
Philadelphia, United States\\
yunchong@alumni.upenn.edu}

\and
\IEEEauthorblockN{6\textsuperscript{th} Bolin Huang}
\IEEEauthorblockA{
\textit{University of Southern California}\\
California, United States\\
bolinhua@usc.edu}
\and
\IEEEauthorblockN{7\textsuperscript{th} Chunhong Yuan}
\IEEEauthorblockA{
\textit{Kazan Federal University}\\
Kazan, Russia\\
ChYuan@kpfu.ru}

\thanks{\textsuperscript{$\dagger$}These authors contributed equally to this work.}
}

\maketitle

\begin{abstract}
Aiming at the problem of difficulty in accurately identifying graphical implicit correlations in multimodal irony detection tasks, this paper proposes a Semantic Irony Recognition Network (SemIRNet). The model contains three main innovations: (1) The ConceptNet knowledge base is introduced for the first time to acquire conceptual knowledge, which enhances the model's common-sense reasoning ability; (2) Two cross-modal semantic similarity detection modules at the word level and sample level are designed to model graphic-textual correlations at different granularities; and (3) A contrastive learning loss function is introduced to optimize the spatial distribution of the sample features, which improves the separability of positive and negative samples. Experiments on a publicly available multimodal irony detection benchmark dataset show that the accuracy and F1 value of this model are improved by 1.64\% and 2.88\% to 88.87\% and 86.33\%, respectively, compared with the existing optimal methods. Further ablation experiments verify the important role of knowledge fusion and semantic similarity detection in improving the model performance.
\end{abstract}

\begin{IEEEkeywords}
Multimodel Learning, Knowledge Fusion, Semantic Similarity, Comparative Learning
\end{IEEEkeywords}

\section{Introduction}
With the rapid development of social media, users generate a large number of multimodal messages containing images and texts. In these messages, ironic expressions are becoming more and more common as a specific linguistic phenomenon. Irony is defined as "saying or writing the opposite of what is actually intended, or speaking with the intention of making others feel stupid or letting them know that you are angry". Irony detection is crucial to the task of sentiment analysis because ironic language often expresses the opposite emotional polarity of its intended meaning compared to its literal meaning \cite{baltrusaitis2018multimodal,agrawal2016analyzing,anagnostopoulos2015features,finn2018learning}.

Early irony language detection mainly utilized textual information. With the increase of multimodal data, more and more studies have started to focus on multimodal irony detection. Existing multimodal irony detection models are mainly based on two approaches: (1) Attention mechanism-based models capture inter-modal inconsistencies by designing different deformation structures of the attention mechanism; and (2) graph neural network-based models establish correlations of multimodal information by constructing cross-modal graph networks. However, these approaches mainly focus on the surface features of images and texts, ignoring the importance of commonsense knowledge for detecting and understanding non-literal expressions. In the field of affective computing, the integration of commonsense knowledge has been shown to help improve the performance of models, as acquiring commonsense knowledge through task-specific dataset learning alone is inherently difficult\cite{anderson2013expressive,andreas2016neural,andrew2013deep}.

\begin{figure}[h!]
    \centering
    \includegraphics[width=0.7\linewidth]{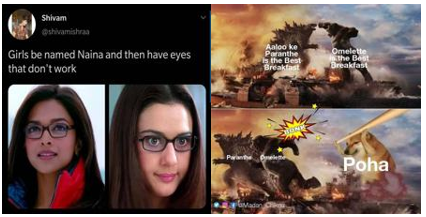}
    \caption{Examples of non-ironic and ironic samples.}
    \label{fig:Examples of non-ironic and ironic samples}
\end{figure}

Based on this, this paper constructs a knowledge fusion model that conforms to the human cognitive style from the perspective of human recognition of ironic language. As shown in Fig. \ref{fig:Examples of non-ironic and ironic samples}. in non-ironic multimodal information, the semantic information in text and images is often explicitly related. On the contrary, in ironic samples, the semantic information in text and images tends to be opposite or implicit, i.e., the image-text information is implicitly correlated. Such implicitly correlated multimodal data need to be identified with some commonsense information, as implicit emotional expressions often require a more conceptual understanding of words in different situations.

(1) The main contributions of this paper are as follows:
A novel Semantic Irony Recognition Network that integrates the ConceptNet knowledge base for multimodal irony detection is proposed. To the best of our knowledge, this is the first model to incorporate conceptual knowledge to enhance irony detection accuracy.

(2) A new multimodal information processing method is designed, featuring word-level and sample-level cross-modal semantic similarity detection modules. These modules assess the semantic consistency of different modalities, enabling the model to extract features relevant to irony detection.

(3) Contrastive learning is introduced to distinguish ironic (positive) and non-ironic (negative) samples. A contrastive learning loss function is employed to refine multimodal feature representation. The proposed model achieves state-of-the-art performance on public benchmark datasets.

The paper is organized as follows: section 2 introduces the related work; section 3 describes the proposed SemIRNet model in detail; section 4 gives the experimental results and analysis; and section 5 concludes the paper.

\section{RELATER WORK}
This section presents related work in two main areas: multimodal learning and multimodal irony detection.
\subsection{Multimodal Learning}
Multimodal learning is a method of content analysis and understanding using multiple modes of information delivery. Since multimedia data usually contains multiple modal information such as image and text, multimodal learning has become the main method for multimedia content analysis. Some representative works in image-text information fusion have emerged in recent years:
(1) The ERNIE-ViLM model utilizes structured knowledge in scene graphs to enable the model to perform fine-grained semantic alignment. 
(2) The VIVO model uses Image-Tag for pre-training to align semantic labels with region features in images. 
(3) The RpBERT model uses multimodal Bidirectional Encoder Representations from Transformers (BERT) for entity recognition tasks, and the proposed relation propagation mechanism allows for better utilization of visual information based on the correlation between images and text.

\subsection{Multimodal Irony Detection}
\subsubsection{Text-based Approach}
Early research on text-based irony detection primarily relied on traditional natural language processing techniques, such as lexical annotation, syntactic analysis, and linguistic feature extraction, to identify ironic expressions. With advancements in deep learning, researchers introduced neural network models to enhance detection performance. For instance, textCNN \cite{rakhlin2016convolutional} employs convolutional neural networks to capture local semantic features of text. The emergence of pre-trained language models further revolutionized irony detection by leveraging large-scale pre-training to achieve robust language understanding. These models also introduced parameter-sharing techniques, reducing model size while maintaining performance, thereby significantly improving text-based irony detection accuracy. However, relying solely on textual features presents limitations. These methods struggle with ironic expressions that require background knowledge, fail to incorporate auxiliary information such as images in social media, and overly depend on specific linguistic patterns, limiting their generalization. These challenges underscore the necessity of developing multimodal irony detection methods.

\subsubsection{Text-based Approach}
With the rapid growth of social media, user-generated content increasingly exhibits multimodal characteristics, combining text and images \cite{li2025enhanced}. Compared to text-only methods, multimodal irony detection emphasizes inter-modal information fusion and interaction to better capture users' expressive intent. Two primary technical approaches have emerged: attention-based methods and graph neural networks (GNNs).

(1)Attention-based methods dynamically establish associations between modalities to capture irony-related features. For example, Pan et al. proposed a cross-modal attention model based on BERT \cite{pan2022probabilistic}, which detects intra-modal and inter-modal semantic inconsistencies. Additionally, researchers have developed end-to-end text-visual fusion models based on Transformers, further improving modality integration.

(2) GNN-based methods construct cross-modal interaction graphs to model ironic features. This graph-based approach is particularly effective in handling the complex relationships between multimodal data \cite{zhao2024towards}.

Despite notable advancements, existing multimodal irony detection approaches face key challenges. They often prioritize surface feature alignment over deep semantic understanding and fail to incorporate common-sense knowledge, leading to poor performance in complex scenarios \cite{cao2024systematic}. Additionally, simplistic modality fusion strategies fail to fully leverage the complementary strengths of different modalities\cite{li2024prototype}. These limitations highlight the need for cognitively inspired multimodal irony detection models that better align with human understanding.

\section{Method}
In this paper, we propose a Semantic Irony Recognition Network called SemIRNet for multimodal irony detection. The model architecture contains the following main components:

(1) Text and image feature extraction module: use pre-trained BERT and ResNet to encode text and image information respectively

(2))ConceptNet-based Knowledge Enhancement Module: Introducing Conceptual Knowledge to Enhance Common Sense Reasoning in Models 

(3)Cross-modal semantic similarity detection module: two levels: word-level and sample-level 

(4)Comparative learning optimization module: improving feature space distribution

\subsection{Technical Details}
For text input, the model utilizes a pre-trained BERT encoder, which separately processes the main text and title text, outputting corresponding feature vectors. For image data, visual features are extracted using ResNet, and an average pooling layer is applied to obtain a fixed-dimensional feature representation.

To enhance semantic understanding, we incorporate the ConceptNet knowledge base, enriching text and image attributes at the conceptual level \cite{dan2024image}. Specifically, ConceptNet is queried to retrieve related concepts and relationships associated with text and image attribute words, expanding semantic information \cite{wei2023breast}. This conceptual information is then encoded into vector form for subsequent semantic similarity computation.

For knowledge enhancement, while simple feature alignment effectively captures explicit modal associations, our framework integrates deeper semantic understanding through ConceptNet. For example, when processing contrasting pairs like "sunny photo" and "bad day", the knowledge enhancement module establishes semantic connections beyond surface-level feature matching by leveraging conceptual relationships from ConceptNet. This approach balances computational efficiency with semantic comprehension \cite{zhang2023denoising}. The knowledge integration process involves querying related concepts, encoding them into vector form, and computing semantic similarities.

To achieve multi-level semantic similarity detection, we design two mechanisms:

(1) Word-level similarity detection: Conceptual representations of text and image attributes are compared using matrix operations, with maximum pooling extracting the most significant semantic associations. This allows the model to capture fine-grained cross-modal correspondences.

(2) Sample-level spatial mapping: A spatial mapping mechanism aligns feature spaces across modalities. Text features (including main text and captions) are concatenated with image features, and a sample covariance matrix is computed to derive a mapping matrix. This transformation projects features into a shared semantic space, facilitating similarity computation.

To further enhance classification performance, we introduce a triad-based contrastive learning mechanism. During training, for each anchor sample, positive samples (from the same category) and negative samples (from different categories) are selected. The model optimizes feature space distribution by minimizing the distance between anchor and positive samples while maximizing the distance between anchor and negative samples.

During training, the model is trained end-to-end using the Adam optimizer, with a learning rate of $1 \times 10^{-5}$ and a batch size of 32. In the inference stage, input data undergo feature extraction, knowledge enhancement, and semantic similarity detection, ultimately producing the irony detection result.

\subsection{Implementation Details}
This section details the key settings of the model implementation. Table  \ref{tab:Key Configuration Parameters of SemIRNet} summarizes the main model configuration parameters:

\begin{table}[htbp]
\footnotesize
\centering
\caption{Key Configuration Parameters of SemIRNet.}
\label{tab:Key Configuration Parameters of SemIRNet}
\begin{tabular}{|c|c|c|}
\hline
\multicolumn{1}{|c|}{Module} 
& \multicolumn{1}{c|}{Parameter} 
& \multicolumn{1}{c|}{Setting} \\
\hline
\multirow{3}{*}{$\text{Text Encoding}$} 
& $\text{Pre-trained Model}$ 
& $\text{BERT-base}$ \\
\cline{2-3}
& $\text{Hidden Dimension}$ 
& $\text{768}$ \\
\cline{2-3}
& $\text{Max Length}$ 
& $\text{128 token}$ \\
\hline

\multirow{3}{*}{$\text{Image Encoding}$}
& $\text{Backbone}$ 
& $\text{ResNet-152}$ \\
\cline{2-3}
& $\text{Input Size}$ 
& $\text{224×224}$ \\
\cline{2-3}
& $\text{Caption Model}$ 
& $\text{MobileNetV3}$ \\
\hline

$\text{Knowledge Integration}$
& $\text{ConceptNet Dimension}$ 
& $\text{300}$ \\
\hline  

\multirow{5}{*}{$\text{Training}$}
& $\text{Optimizer}$ 
& $\text{Adam}$ \\
\cline{2-3}
& $\text{Learning Rate}$ 
& $1 \mathrm{e}{-5}$ \\
\cline{2-3}
& $\text{Batch Size}$ 
& $\text{32}$ \\
\cline{2-3}
& $\text{Contrastive Margin}$ 
& $\text{0.5}$ \\
\cline{2-3}
& $\text{Loss Weight $\lambda$}$ 
& $\text{0.1}$ \\
\hline
\end{tabular}
\end{table}

In the process of model training, we adopt a series of strategies to ensure the stability of performance and generalization ability. Firstly, in the data preprocessing stage, the text data are randomly masked with the mask ratio set to 15\%, and the image data are enhanced with random cropping, horizontal flipping and normalization. These data enhancement techniques effectively improve the robustness of the model.

The training adopts a staged strategy: firstly, the text and image encoders are pre-trained to obtain the basic feature representation capability, then the ConceptNet knowledge vectors are loaded for knowledge enhancement, and finally, the end-to-end model training is carried out. During the training process, we use the performance of the validation set as the early stopping criterion, and dynamically adjust the hyperparameters accordingly. 

The whole training process is carried out on a single RTX 4060 graphics card. Through the optimization of the above implementation details, the model is able to achieve stable and excellent performance. The experimental results show that the choice of these parameter configurations and training strategies is crucial to realize the potential of the model in the multimodal irony detection task.

\section{Experiments}
\subsection{Main Results}
We evaluated the performance of SemIRNet on a publicly available multimodal irony detection benchmark dataset. Fig. \ref{fig:Visualization of model performance comparison on Dataset-1}. and Table \ref{tab:Performance} show the experimental results:

\begin{figure}[h!]
    \centering
    \includegraphics[width=0.8\linewidth]{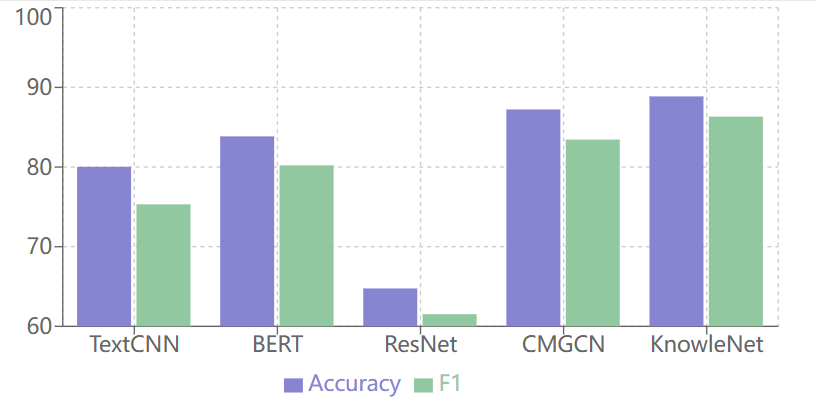}
    \caption{Visualization of model performance comparison on Dataset-1.}
    \label{fig:Visualization of model performance comparison on Dataset-1}
\end{figure}

\begin{table}[htbp]
\footnotesize
\centering
\caption{Performance Comparison on Dataset-1 (\%).}
\label{tab:Performance}
\begin{tabular}{|c|c|c|c|c|}
\hline
\multicolumn{1}{|c|}{Model} 
& \multicolumn{1}{c|}{Accuracy} 
& \multicolumn{1}{c|}{Precision} 
& \multicolumn{1}{c|}{Recall} 
& \multicolumn{1}{c|}{F1-score} \\
\hline

$ \text{TextCNN} $ 
& $\text{80.03}$ 
& $\text{74.29}$ 
& $\text{76.39}$ 
& $\text{75.32}$ \\
\hline
$ \text{BERT} $ 
& $\text{83.85}$ 
& $\text{78.72}$ 
& $\text{82.27}$ 
& $\text{80.22}$ \\
\hline
$ \text{ResNet} $ 
& $\text{64.76}$ 
& $\text{54.41}$ 
& $\text{70.80}$ 
& $\text{61.53}$ \\
\hline
$ \text{ViT} $ 
& $\text{67.83}$ 
& $\text{57.93}$ 
& $\text{70.07}$ 
& $\text{63.43}$ \\
\hline
$ \text{HFM} $ 
& $\text{83.44}$ 
& $\text{76.57}$ 
& $\text{84.15}$ 
& $\text{80.18}$ \\
\hline
$ \text{D\&R Net} $ 
& $\text{84.02}$ 
& $\text{77.97}$ 
& $\text{83.42}$ 
& $\text{80.60}$ \\
\hline
$ \text{CMGCN} $ 
& $\text{87.23}$ 
& $\text{-}$ 
& $\text{-}$ 
& $\text{83.45}$ \\
\hline
$ \text{SemIRNet} $ 
& $\textbf{88.87}$ 
& $\textbf{88.59}$ 
& $\textbf{84.18}$ 
& $\textbf{86.33}$ \\
\hline
\end{tabular}
\end{table}

The following key findings can be observed from the experimental results:
Among the unimodal methods, the text-based methods generally outperform the image-based methods. Among them, the BERT model shows the best unimodal performance, reaching 83.85\% accuracy and 80.22\% F1 value. This indicates that textual information is more discriminative than visual information in irony detection tasks.
Multimodal methods have significantly improved their performance by fusing text and image information. Earlier multimodal methods such as Hierarchical Fusion Model(HFM) and Detection and Recognition Network(D\&R) Net have demonstrated better performance than unimodal methods. The latest CMGCN model based on graph neural network further improves the accuracy to 87.23\%.
Our proposed SemIRNet model achieves optimal performance on all evaluation metrics by introducing knowledge enhancement and multi-level semantic similarity detection. Specifically:

(1) Accuracy of 88.87 per cent, an improvement of 1.64 per cent compared to CMGCN

(2) F1 value reached 86.33\%, an improvement of 2.88\% compared to CMGCN

(3) Achieved a balanced improvement in both precision and recall rates

These results confirm the effectiveness of knowledge fusion and semantic alignment in improving the performance of multimodal irony detection, especially when dealing with complex samples that require deep semantic understanding, our approach shows significant advantages.

\subsection{Ablation Study}
In order to verify the effectiveness of each key component of the model, we conducted detailed ablation experiments. Fig. \ref{fig:Visualization}. and Table \ref{tab:Ablation} demonstrate the experimental results:

\begin{figure}[h!]
    \centering
    \includegraphics[width=0.8\linewidth]{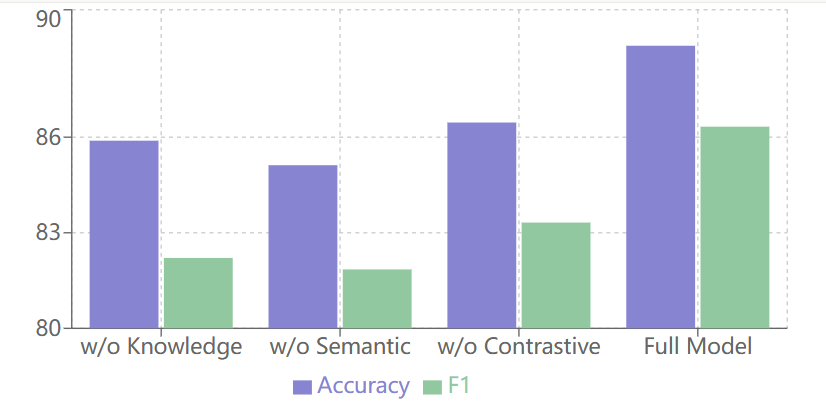}
    \caption{Visualization of ablation experiment results. }
    \label{fig:Visualization}
\end{figure}

\begin{table}[htbp]
\scriptsize
\centering
\caption{Ablation Study Results on Dataset-1 (\%).}
\label{tab:Ablation}
\begin{tabular}{|c|c|c|c|c|c|c|}
\hline
\makecell{Model \\ Variant} & Accuracy & $\Delta$Acc & F1-score & $\Delta$F1 & \makecell{Macro \\ -F1} & \makecell{$\Delta$Macro \\ -F1} \\
\hline

\makecell{$\textbf{SemIRNet}$} 
& $\textbf{88.87}$ 
& $\text{-}$ 
& $\textbf{86.33}$ 
& $\text{-}$ 
& $\textbf{88.51}$
& $\text{-}$ \\
\hline

\makecell{$\text{w/o}$ \\ $\text{Knowledge}$} 
& $\text{85.89}$ 
& $\text{-2.98}$ 
& $\text{82.21}$ 
& $\text{-4.12}$ 
& $\text{84.58}$
& $\text{-3.93}$ \\
\hline

\makecell{$\text{w/o}$ \\ Semantic} 
& $\text{85.12}$ 
& $\text{-3.75}$ 
& $\text{81.85}$ 
& $\text{-4.48}$ 
& $\text{83.38}$
& $\text{-5.13}$ \\
\hline
\makecell{$\text{w/o}$  \\ Contrastive} 
& $\text{86.46}$ 
& $\text{-2.41}$ 
& $\text{83.32}$ 
& $\text{-3.01}$ 
& $\text{85.63}$
& $\text{-2.88}$ \\
\hline
\end{tabular}
\end{table}

(1) Knowledge Enhancement Module: Removing the Knowledge Enhancement Module (w/o Knowledge) causes a 2.98\% decrease in accuracy, and a more significant decrease in F1 value (4.12\%), which suggests that knowledge fusion is crucial to the model's discriminative ability, which validates the need to introduce ConceptNet for conceptual-level semantic enhancement.

(2) Semantic Similarity Detection: The removal of Semantic Similarity Detection (w/o Semantic) causes the largest performance degradation (-3.75\% accuracy), and Macro-F1 decreases by 5.13\%, which indicates that this module is particularly important for dealing with data imbalance. The results confirm the key role of multilevel Semantic Similarity Detection in capturing inter-modal relationships.

(3) Contrastive Learning Optimization: Removing the contrastive learning loss (w/o Contrastive) decreases the accuracy by 2.41\%. The performance degradation is relatively small but still significant, indicating that contrastive learning does help the model to learn a more discriminative feature representation.

These experimental results clearly show that each component of the model contributes significantly to the final performance. In particular, the importance of the semantic similarity detection module is most prominent, which is in line with our original design intention of emphasizing deep semantic understanding. Also, the synergistic effect of knowledge enhancement and comparison learning is shown to be a key factor in improving the model's performance.

The experimental results demonstrate that our modal fusion strategy effectively balances performance and computational complexity. Ablation studies validate the contribution of each component and identify opportunities for future improvement. Looking ahead, more sophisticated fusion approaches could be explored to further enhance the model's capabilities. Potential extensions include adaptive attention mechanisms that dynamically adjust based on modal consistency \cite{panda2021adamml}, multi-level semantic alignment frameworks \cite{xu2024hka}, and cognitive-inspired information integration patterns \cite{chen2022decade}. These advanced techniques could improve the model's ability to interpret complex ironic expressions while maintaining computational efficiency. The performance gains observed in our experiments suggest that such enhancements could lead to meaningful improvements in multimodal irony detection.

\subsection{Qualitative Analysis}

(1)Performance Advantage: SemIRNet achieves optimal performance on several datasets, especially on Dataset-1 where the accuracy and F1 value are improved by 1.64\% and 2.88\%, respectively, compared to the best baseline CMGCN. This confirms the effectiveness of our proposed knowledge fusion-based approach on the multimodal irony detection task.

(2)Module contribution: The ablation experiments clearly demonstrate the importance of each technology module: the semantic similarity detection module contributes the most ($\Delta$Acc: -3.75\%), the knowledge enhancement module is the second most important ($\Delta$Acc: -2.98\%), and the contrastive learning optimisation also plays a significant role ($\Delta$
Acc: -2.41\%).  

(3)Methodological innovations: The experimental results validate our three main innovations: the introduction of ConceptNet for knowledge enhancement indeed improves the semantic comprehension of the model, multi-level semantic similarity detection effectively captures the complex relationships between modalities, and contrastive learning optimization improves the distribution structure of the feature space.

While our experimental results show that surface feature matching can achieve satisfactory accuracy (86.33\% F1-score) with relatively low computational complexity, more sophisticated fusion schemes can be explored to further enhance the model's performance. Advanced approaches such as hierarchical attention fusion \cite{tao2024hierarchical}, dynamic modal alignment \cite{nadeem2024cad}, and cognitive-inspired integration mechanisms \cite{chen2024cognitive} could potentially improve the model's ability to capture implicit semantic relationships.

\section{Conclusion}
In this paper, we propose Semantic Irony Recognition Network (SemIRNet) to address the challenge of accurately identifying implicit associations in multimodal irony detection. Experiments on multiple public datasets demonstrate that our model improves accuracy and F1-score by 1.64\% and 2.88\%, respectively, compared to existing state-of-the-art methods. Ablation studies further validate the effectiveness of each module, particularly highlighting the significant contributions of semantic similarity detection and knowledge enhancement to overall performance.

For future work, we aim to explore several promising directions to further enhance the model's capabilities. First, we plan to investigate advanced fusion mechanisms, such as hierarchical attention networks \cite{baltrusaitis2018multimodal} and dynamic modal alignment \cite{agrawal2016analyzing}, to improve the model's ability to capture implicit semantic relationships between visual and textual content. Additionally, we intend to develop lightweight yet effective knowledge enhancement techniques \cite{andreas2016neural} that can enrich semantic understanding while maintaining computational efficiency.

Another key direction is the investigation of cross-domain adaptation methods, which could significantly improve the model's generalization across different social media platforms and content types. These extensions are expected to lead to more robust and efficient multimodal irony detection systems, addressing current limitations in handling complex ironic expressions and context-dependent cases, while ensuring practical computational efficiency for real-world applications.

\bibliographystyle{IEEEtran}
\bibliography{ref}

\begin{thebibliography}{10}
\providecommand{\url}[1]{#1}
\csname url@samestyle\endcsname
\providecommand{\newblock}{\relax}
\providecommand{\bibinfo}[2]{#2}
\providecommand{\BIBentrySTDinterwordspacing}{\spaceskip=0pt\relax}
\providecommand{\BIBentryALTinterwordstretchfactor}{4}
\providecommand{\BIBentryALTinterwordspacing}{\spaceskip=\fontdimen2\font plus
\BIBentryALTinterwordstretchfactor\fontdimen3\font minus \fontdimen4\font\relax}
\providecommand{\BIBforeignlanguage}[2]{{%
\expandafter\ifx\csname l@#1\endcsname\relax
\typeout{** WARNING: IEEEtran.bst: No hyphenation pattern has been}%
\typeout{** loaded for the language `#1'. Using the pattern for}%
\typeout{** the default language instead.}%
\else
\language=\csname l@#1\endcsname
\fi
#2}}
\providecommand{\BIBdecl}{\relax}
\BIBdecl

\bibitem{baltrusaitis2018multimodal}
T.~Baltru{\v{s}}aitis, C.~Ahuja, and L.-P. Morency, ``Multimodal machine learning: A survey and taxonomy,'' \emph{IEEE Transactions on Pattern Analysis and Machine Intelligence}, vol.~41, no.~2, pp. 423--443, 2018.

\bibitem{agrawal2016analyzing}
A.~Agrawal, D.~Batra, and D.~Parikh, ``Analyzing the behavior of visual question answering models,'' \emph{arXiv preprint arXiv:1606.07356}, 2016.

\bibitem{anagnostopoulos2015features}
C.~Anagnostopoulos, T.~Iliou, and I.~Giannoukos, ``Features and classifiers for emotion recognition from speech: a survey from 2000 to 2011,'' \emph{Artificial Intelligence Review}, vol.~43, pp. 155--177, 2015.

\bibitem{finn2018learning}
C.~Finn, ``Learning to learn with gradients,'' Master's thesis, University of California, Berkeley, 2018.

\bibitem{anderson2013expressive}
R.~Anderson, B.~Stenger, V.~Wan \emph{et~al.}, ``Expressive visual text-to-speech using active appearance models,'' in \emph{Proceedings of the IEEE Conference on Computer Vision and Pattern Recognition}.\hskip 1em plus 0.5em minus 0.4em\relax IEEE, 2013, pp. 3382--3389.

\bibitem{andreas2016neural}
J.~Andreas, M.~Rohrbach, T.~Darrell \emph{et~al.}, ``Neural module networks,'' in \emph{Proceedings of the IEEE Conference on Computer Vision and Pattern Recognition}.\hskip 1em plus 0.5em minus 0.4em\relax IEEE, 2016, pp. 39--48.

\bibitem{andrew2013deep}
G.~Andrew, R.~Arora, J.~Bilmes \emph{et~al.}, ``Deep canonical correlation analysis,'' in \emph{International Conference on Machine Learning}.\hskip 1em plus 0.5em minus 0.4em\relax PMLR, 2013, pp. 1247--1255.

\bibitem{rakhlin2016convolutional}
A.~Rakhlin, ``Convolutional neural networks for sentence classification,'' \emph{GitHub}, vol.~6, p.~25, 2016.

\bibitem{li2025enhanced}
L.~Li, R.~Wang, M.~Zou, F.~Guo, and Y.~Ren, ``Enhanced resnet-50 for garbage classification: Feature fusion and depth-separable convolutions,'' \emph{PloS one}, vol.~20, no.~1, p. e0317999, 2025.

\bibitem{pan2022probabilistic}
M.~Pan, J.~Wang, J.-X. Huang \emph{et~al.}, ``A probabilistic framework for integrating sentence-level semantics via bert into pseudo-relevance feedback,'' \emph{Information Processing \& Management}, vol.~59, no.~1, p. 102734, 2022.

\bibitem{zhao2024towards}
G.~Zhao, P.~Li, Z.~Zhang, F.~Guo, X.~Huang, W.~Xu, J.~Wang, and J.~Chen, ``Towards sar automatic target recognition: Multi-category sar image classification based on light weight vision transformer,'' in \emph{2024 21st Annual International Conference on Privacy, Security and Trust (PST)}.\hskip 1em plus 0.5em minus 0.4em\relax IEEE, 2024, pp. 1--6.

\bibitem{cao2024systematic}
Y.~Cao, J.~Dai, Z.~Wang, Y.~Zhang, X.~Shen, Y.~Liu, and Y.~Tian, ``Systematic review: Text processing algorithms in machine learning and deep learning for mental health detection on social media,'' \emph{arXiv preprint arXiv:2410.16204}, 2024.

\bibitem{li2024prototype}
L.~Li, Z.~Li, F.~Guo, H.~Yang, J.~Wei, and Z.~Yang, ``Prototype comparison convolutional networks for one-shot segmentation,'' \emph{IEEE Access}, 2024.

\bibitem{dan2024image}
H.-C. Dan, Z.~Huang, B.~Lu, and M.~Li, ``Image-driven prediction system: Automatic extraction of aggregate gradation of pavement core samples integrating deep learning and interactive image processing framework,'' \emph{Construction and Building Materials}, vol. 453, p. 139056, 2024.

\bibitem{wei2023breast}
Y.~Wei, D.~Zhang, M.~Gao, Y.~Tian, Y.~He, B.~Huang, and C.~Zheng, ``Breast cancer prediction based on machine learning,'' \emph{Journal of Software Engineering and Applications}, vol.~16, pp. 348--360, 2023.

\bibitem{zhang2023denoising}
D.~Zhang, F.~Zhou, Y.~Wei, X.~Yang, and Y.~Gu, ``Unleashing the power of self-supervised image denoising: A comprehensive review,'' \emph{arXiv preprint arXiv:2308.00247}, 2023.

\bibitem{panda2021adamml}
R.~Panda, C.~Chen, Q.~Fan \emph{et~al.}, ``Adamml: Adaptive multi-modal learning for efficient video recognition,'' in \emph{Proceedings of the IEEE/CVF International Conference on Computer Vision}, 2021, pp. 7576--7585.

\bibitem{xu2024hka}
Y.~Xu, Y.~Li, M.~Xu \emph{et~al.}, ``Hka: A hierarchical knowledge alignment framework for multimodal knowledge graph completion,'' \emph{ACM Transactions on Multimedia Computing, Communications, and Applications}, vol.~20, no.~8, pp. 1--19, 2024.

\bibitem{chen2022decade}
X.~Chen, H.~Xie, G.~Cheng \emph{et~al.}, ``A decade of sentic computing: topic modeling and bibliometric analysis,'' \emph{Cognitive Computation}, vol.~14, no.~1, pp. 24--47, 2022.

\bibitem{tao2024hierarchical}
H.~Tao and Q.~Duan, ``Hierarchical attention network with progressive feature fusion for facial expression recognition,'' \emph{Neural Networks}, vol. 170, pp. 337--348, 2024.

\bibitem{nadeem2024cad}
A.~Nadeem, A.~Hilton, R.~Dawes \emph{et~al.}, ``Cad-contextual multi-modal alignment for dynamic avqa,'' in \emph{Proceedings of the IEEE/CVF Winter Conference on Applications of Computer Vision}, 2024, pp. 7251--7263.

\bibitem{chen2024cognitive}
X.~Chen, H.~Xie, S.~Qin \emph{et~al.}, ``Cognitive-inspired deep learning models for aspect-based sentiment analysis: A retrospective overview and bibliometric analysis,'' \emph{Cognitive Computation}, pp. 1--39, 2024.

\end{thebibliography}

\end{document}